\documentclass[sigconf,pbalance]{acmart}

\usepackage{subcaption}
\usepackage[linesnumbered,ruled,vlined]{algorithm2e}
\usepackage{colortbl}
\usepackage{makecell}

\SetKwComment{Comment}{/* }{ */ \hfill}
\SetCommentSty{commentstyle}
\SetKwInput{Require}{Require}
\SetKwInput{Ensure}{Ensure}
\definecolor{TableGroupGray}{RGB}{238,238,238}
\definecolor{TableMethodBlue}{RGB}{229,238,250}

\copyrightyear{2027}
\acmYear{2027}
\setcopyright{cc}
\setcctype{by}
\acmConference[WSDM '27]{Proceedings of the Twentieth ACM International Conference on Web Search and Data Mining}{February 15--19, 2027}{Hong Kong, Hong Kong}
\acmBooktitle{Proceedings of the Twentieth ACM International Conference on Web Search and Data Mining (WSDM '27), February 15--19, 2027, Hong Kong, Hong Kong}
\acmDOI{10.1145/3862322.3863902}
\acmISBN{979-8-4007-3117-4/2027/02}
\begin{document}

\title[ERPO: Token-Level Entropy-Regulated Policy Optimization]{ERPO: Token-Level Entropy-Regulated Policy Optimization for Large Reasoning Models}

\author{Song Yu}
\orcid{0009-0008-8525-6617}
\affiliation{%
  \department{School of Computer and Information Science}
  \institution{Southwest University}
  \city{Chongqing}
  \country{China}}
\email{yusong0929@email.swu.edu.cn}

\author{Li Li}
\authornote{Corresponding author: Li Li, Southwest University, Chongqing 400715, China.}
\orcid{0000-0003-4818-8770}
\affiliation{%
  \department{School of Computer and Information Science}
  \institution{Southwest University}
  \city{Chongqing}
  \country{China}}
\email{lily@swu.edu.cn}

\author{Wenwen Zhao}
\orcid{0009-0004-4891-2292}
\affiliation{%
  \department{School of Computer and Information Science}
  \institution{Southwest University}
  \city{Chongqing}
  \country{China}}
\email{zhaoww71@email.swu.edu.cn}

\author{Zhisheng Yang}
\orcid{0000-0003-2458-5114}
\affiliation{%
  \department{School of Computer and Information Science}
  \institution{Southwest University}
  \city{Chongqing}
  \country{China}}
\email{yzsheng@email.swu.edu.cn}

\renewcommand{\shortauthors}{Song Yu, Li Li, Wenwen Zhao, and Zhisheng Yang}

\begin{abstract}

Reinforcement learning from verifiable rewards has substantially improved the reasoning capabilities of large language models. Group Relative Policy Optimization (GRPO), however, broadcasts one sequence-level advantage to every token in a response and therefore cannot distinguish token positions with different uncertainty and learning dynamics. We empirically observe that perturbations targeting localized high-entropy regions are more damaging than perturbations targeting low-entropy regions.
Motivated by this observation, we propose \textbf{ERPO} (Entropy-Regulated Policy Optimization), a token-level credit-assignment method that uses a bounded entropy gate to emphasize relatively uncertain tokens, normalizes reference-relative policy changes within relative-position buckets, and anchors the resulting token signal to the sign of the sequence-level advantage. ERPO requires neither a learned process-reward model nor additional supervision and can be implemented on top of the GRPO training pipeline. Experiments on four mathematical reasoning benchmarks across three Qwen2.5 model scales evaluate accuracy, response length, output validity, and component contributions. ERPO yields its clearest gains over matched GRPO at the 3B scale and remains competitive at 1.5B and 7B, while avoiding the late-stage response-length and truncation degradation observed in the matched 3B normalization runs.

\end{abstract}




\maketitle

\section{INTRODUCTION}

\begin{figure*}
	\centering
	% 左侧子图 (a)
	\begin{subfigure}[b]{0.65\textwidth}
		\centering
		\includegraphics[width=\textwidth]{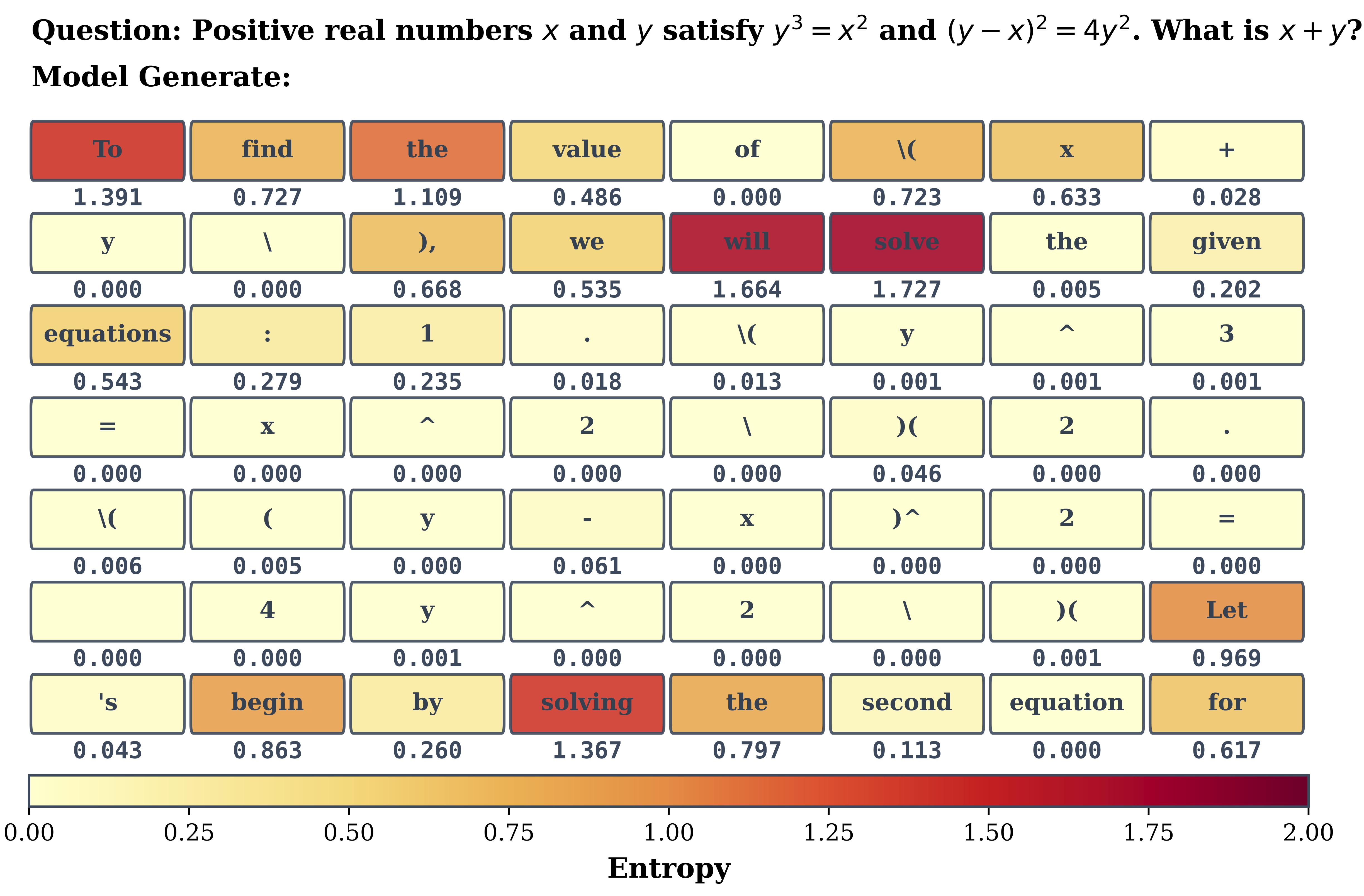}
		\caption{Token-level entropy distribution.}
		\label{fig:entropy_map}
	\end{subfigure}
	\hfill % 在两个子图之间插入弹性间距
	% 右侧子图 (b)
	\begin{subfigure}[b]{0.32\textwidth}
		\centering
		\includegraphics[width=\textwidth]{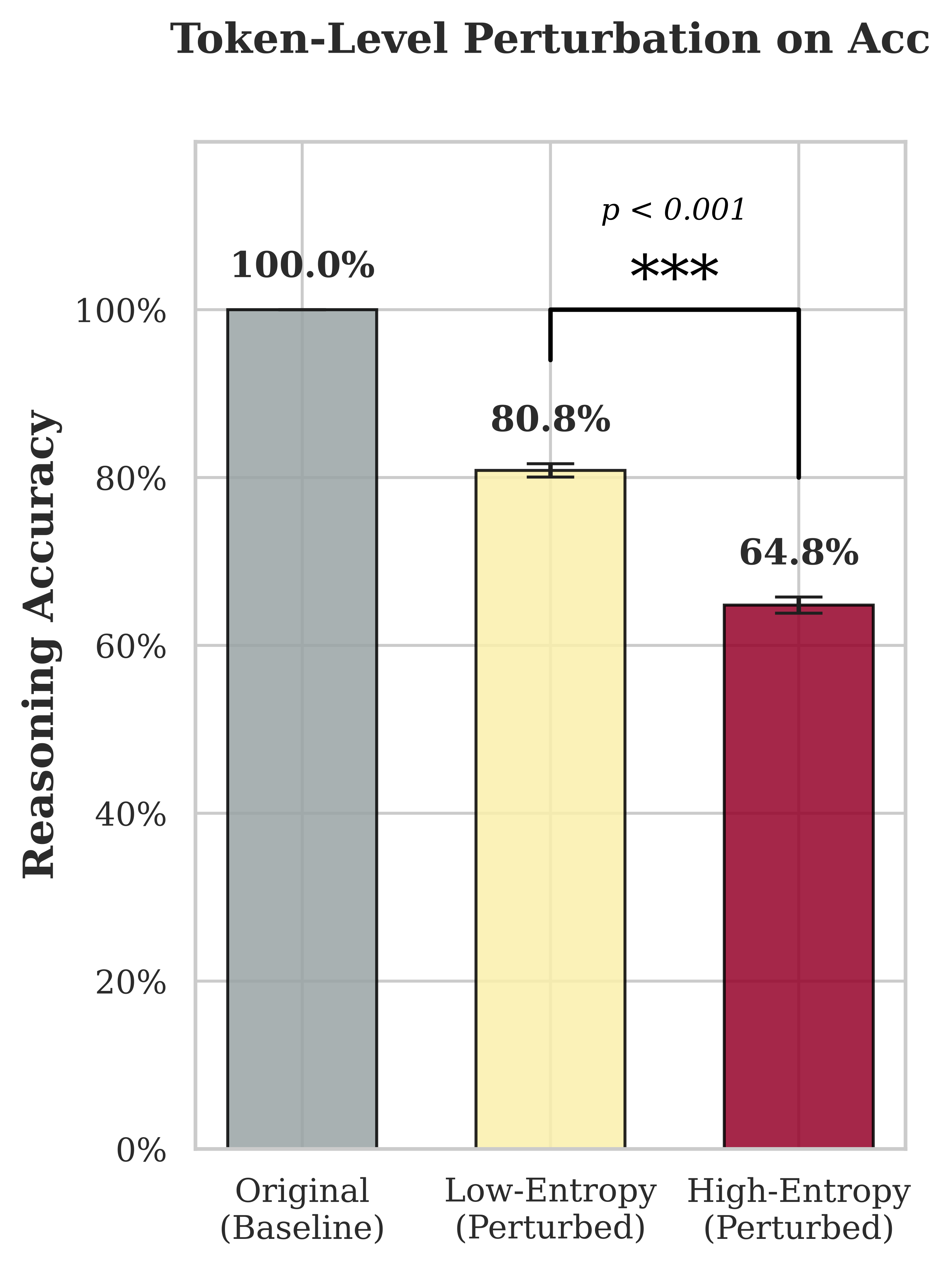}
		\caption{Perturbation sensitivity by entropy.}
		\label{fig:causal_bar}
	\end{subfigure}
	
	\caption{\textnormal{\textbf{Token-level entropy and perturbation sensitivity for Qwen2.5-3B.} (a) Token entropy along one response to an AMC23 problem. (b) For 50 MATH problems that the model initially solved, we ranked tokens by entropy within each response. In each trial, we replaced one token from the top or bottom 5\% with a random vocabulary token and regenerated the continuation. Final-answer accuracy was lower for high-entropy perturbations (2,460 trials per condition, $p<0.001$ in a trial-level two-sample $t$-test).}}
	\label{fig:motivation_main}
	\Description{A heatmap colors each generated token by entropy. The words will and solve have high entropy, while most symbols in the equations have entropy near zero. The adjacent bars show 100 percent accuracy for the original responses, 80.8 percent after low-entropy perturbations, and 64.8 percent after high-entropy perturbations.}
\end{figure*}

Large language models (LLMs) increasingly serve as generative components in search, recommendation, and web-agent systems \citep{sun2023chatgpt,hou2024large,zhou2024webarena}, where post-training must balance capability with training and inference cost. Reinforcement learning has become a central post-training paradigm \citep{ouyang2022training}, and systems such as OpenAI o1 \citep{openai2024learning} and DeepSeek-R1 \citep{guo2025deepseek} demonstrate strong chain-of-thought (CoT) reasoning. A key ingredient is reinforcement learning from verifiable rewards (RLVR) \citep{lamberttulu}, which can use programmatic outcome checks and thereby avoid a learned outcome reward model. Group Relative Policy Optimization (GRPO) \citep{shao2024deepseekmath} further removes the learned critic and estimates advantages by comparing multiple responses to the same prompt, reducing the memory cost of policy optimization.

This setting motivates a common post-training question for long-form generative policies: when supervision is available only at the sequence or task-outcome level, how should credit be distributed across individual tokens? We study this question in mathematical RLVR, where exact-answer verification provides a controlled setting for isolating credit assignment from reward-model error.

Despite the success of GRPO, its credit-assignment rule has an important limitation: every token in a response receives the same sequence-level advantage. This broadcasting rule does not assert that all tokens are equally important, but it cannot represent differences among token positions. In CoT reasoning, predictive uncertainty is heterogeneous: distributions can be diffuse near branch points and concentrated during routine derivations \citep{wangbeyond}. Treating these positions identically may dilute updates at uncertain decisions and spend equal optimization weight on predictable tokens. Existing adjustments to sequence rewards, clipping and KL control, or response-length normalization do not by themselves resolve this token-level distinction \citep{schulman2017proximal,liu2025understanding,yu2025dapo}.

To examine this heterogeneity, we visualize token entropy along inference trajectories and identify localized high-entropy regions using a within-trajectory ranking. As shown in Figure \ref{fig:entropy_map}, these regions contrast with lower-entropy spans associated with more predictable continuation. We treat the ranking only as a measure of relative predictive uncertainty; a high-entropy token need not correspond to a semantic branch point or a uniquely important reasoning step.

In a separate perturbation study on 50 initially solved MATH problems (Figure~\ref{fig:causal_bar}), replacing a high-entropy token reduced final-answer accuracy more than replacing a low-entropy token. This sensitivity gap supports using entropy as a token-level prioritization signal. Because perturbation can also alter downstream contexts, we treat this result as motivational evidence rather than a causal identification result.

Based on this evidence, we propose \textbf{Entropy-Regulated Policy Optimization (ERPO)}. ERPO combines the verifiable sequence outcome with a stop-gradient token-level signal derived from policy change relative to a frozen reference model. Entropy controls where this auxiliary signal receives more weight, relative-position buckets compare tokens at similar stages of differently sized responses, and the sign of the group advantage anchors the signal to the observed outcome. The resulting update reallocates credit within sampled trajectories without training a separate process-reward model.

In summary, our main contributions are as follows:
\begin{itemize}
	\item We provide a perturbation-based analysis showing that response regions selected by within-trajectory entropy ranking exhibit different sensitivity to token replacement.
	\item We introduce ERPO, which differs from sequence-level GRPO by combining entropy gating, relative-position bucket normalization, and outcome anchoring to construct token-level advantages without a learned process-reward model.
	\item We evaluate ERPO at three model scales on four mathematical reasoning benchmarks. ERPO yields its clearest accuracy gains at the 3B scale and maintains more stable response length and output validity than the matched 3B normalization variants, with component ablations and a matched training-cost analysis characterizing the associated trade-offs.
\end{itemize}

\section{Related Work}
\subsection{Sequence-Level RLVR Optimization}
GRPO \citep{shao2024deepseekmath} estimates relative advantages from multiple responses to the same prompt and broadcasts each response advantage to all of its tokens. Subsequent work modifies different parts of this sequence-level interface. GMPO \citep{zhao2025geometric} changes loss aggregation, BAPO \citep{xi2025bapo} adapts clipping for positive and negative advantages, and MAPO \citep{huang2025mapo} modifies group advantages according to trajectory certainty. Dr.~GRPO \citep{liu2025understanding} and DAPO \citep{yu2025dapo} instead address reward- and length-normalization biases through fixed-length and active-token loss normalizers, respectively. 

Reward shaping and multi-objective methods alter the scalar assigned to a complete response. S-GRPO \citep{citation-0} encourages early exit through a decaying sequence reward, while AW-GRPO \citep{ichihara2025auto} and dynamic reward weighting \citep{lu2025learning} combine multiple sequence-level objectives. CoRLHF \citep{liu2025corlhf} jointly updates the policy and a learned reward model. These approaches can change which responses are preferred, whereas ERPO addresses where credit is allocated inside a response and does not train a reward model.

\subsection{Token-Level Credit Assignment}
Annealed decoding varies sampling intensity \citep{yang2025let}, Knapsack RL allocates training effort by task difficulty \citep{li2025knapsack}, and EEPO uses temporary forgetting to encourage exploration \citep{chen2025eepo}. The 80/20 approach keeps policy gradients only for the top 20\% of tokens by entropy in each batch \citep{wangbeyond}. SIREN applies entropy regularization to selected states and actions \citep{jiang2025rethinking}. Clip-Cov and KL-Cov restrict updates for high-covariance tokens through clipping and a KL penalty, respectively \citep{cui2025entropy}. GTPO and GRPO-S use entropy to redistribute rewards at the token and sequence levels, respectively \citep{tan2026gtpo}. HAPO uses entropy and importance ratios to redistribute advantages and adapts sampling temperature and clipping to token entropy \citep{liu2026hapo}. Cheng et al.~\citep{cheng2026exploration} add a capped, detached entropy term to the advantage. In ERPO, entropy scales a separate correction whose sign also depends on the sampled token and the response outcome.

Process supervision can draw on step labels or a learned verifier \citep{setlur2025rewarding}. Other sources of credit include planning structure \citep{dou2025plan}, attention patterns \citep{li2025attention}, and shared rollout trees \citep{li2025treepo}. PRIME trains an implicit process reward model online from rollouts and outcome labels \citep{cui2026prime}. KTAE estimates token importance from occurrence statistics in correct and incorrect rollouts \citep{sun2025ktae}. SPRO and ERPO both derive a token signal from a policy-to-reference log ratio. SPRO accumulates these log ratios over the prefix and subtracts the mean among responses that reach the same token position \citep{fei2025spro}. ERPO normalizes each sampled token's log ratio within a relative-position bucket, weights it by an entropy gate, and orients it with the sign of the response's group advantage.

Systems such as Seer \citep{qin2025seer} optimize rollout and KV-cache efficiency. We compare the end-to-end training costs of ERPO and GRPO under matched 3B training settings.

\section{Preliminaries}
We introduce notation for autoregressive generation and then summarize RLVR and the matched GRPO objective used in our experiments.

\textbf{LLMs.} Given a prompt $x$, an LLM $\pi_{\theta}$ sequentially generates a $T$-token response $o=(o_1,\ldots,o_T)$ \citep{brown2020language}:
\begin{equation}
	\pi_\theta(o\mid x) = \prod_{t=1}^T \pi_\theta(o_t\mid x,o_{<t}).
\end{equation}

\textbf{RLVR.}
RLVR \citep{lamberttulu} uses outcome signals whose correctness can be checked by an explicit verification procedure. Unlike reinforcement learning from human feedback (RLHF) \citep{ouyang2022training}, it can use programmatic or rule-based checks, such as executing code, matching the final answer to a mathematical problem, or verifying an output format, without training a learned outcome reward model. The experiments in this paper use a deterministic final-answer verifier.

Consider a dataset $\mathcal{D}=\{(x,y)\}$ where $x$ is the prompt and $y$ is the reference answer. The optimization objective in RLVR is to maximize the expected reward:

\begin{equation}
	\mathcal{J}(\theta) = \mathbb{E}_{(x,y) \sim \mathcal{D}} \left[ \mathbb{E}_{o \sim \pi_\theta(\cdot\mid x)} [R(o,y)] \right],
\end{equation}

\noindent where $R(o,y)$ compares the generated response $o$ against the reference answer $y$ through the task's verification procedure.

\textbf{GRPO.}
GRPO \citep{shao2024deepseekmath} is more memory efficient than Proximal Policy Optimization (PPO) \citep{schulman2017proximal} because it estimates advantages through group-based response sampling without a learned value network. For each prompt--reference pair $(x,y)$, GRPO samples $G$ responses $\{o_1,o_2,\ldots,o_G\}$ from the behavior policy $\pi_{\theta_{\mathrm{old}}}$, where $|o_i|$ denotes the number of valid completion tokens in response $o_i$. The verifier assigns reward $r_i=R(o_i,y)$.

GRPO first standardizes rewards within the prompt group and then broadcasts each response-level advantage to all of its valid tokens:

\begin{equation}
	\begin{aligned}
	\hat{A}_i^{\mathrm{group}}
	&=\frac{r_i-\operatorname{mean}(\mathbf{r})}
	{\operatorname{std}(\mathbf{r})+10^{-4}},\\
	\hat{A}_{i,t}&=\hat{A}_i^{\mathrm{group}},
	\qquad t=1,\ldots,|o_i|,
	\end{aligned}
\end{equation}

\noindent where $\mathbf{r}=[r_1,r_2,\ldots,r_G]$, and $\operatorname{std}$ denotes the sample standard deviation of the $G$ rewards. This normalization compares responses within the same prompt group. Broadcasting $\hat{A}_i^{\mathrm{group}}$ to every token is an estimator design choice; it does not establish that every token contributes equally to the outcome.

To limit divergence from the reference policy, GRPO uses a non-negative sampled-token KL estimator \citep{shao2024deepseekmath}. Let $u_{i,t}$ denote the reference-to-current probability ratio for the sampled token. Then

\begin{equation}
	\begin{aligned}
	u_{i,t}
	&=\frac{\pi_{\text{ref}}(o_{i,t}\mid x,o_{i,<t})}
	{\pi_\theta(o_{i,t}\mid x,o_{i,<t})},\\
	\widehat{D}_{\mathrm{KL},i,t}
	&=u_{i,t}-\log u_{i,t}-1.
	\end{aligned}
\end{equation}

At a fixed prefix, the expectation of this estimator under current-policy sampling equals $D_{\mathrm{KL}}(\pi_\theta\|\pi_{\mathrm{ref}})$. Differentiating the penalty with the sampled tokens held fixed generally gives a biased estimate of the gradient of that divergence \citep{shahcomedy}. ERPO and the matched GRPO baseline use the same penalty.

The implementation used in our experiments combines a clipped surrogate objective with a KL penalty and normalizes each response by its own number of valid completion tokens:

\begin{equation}
	\begin{aligned}
	\mathcal{L}_{\text{GRPO}}(\theta)
	={}&-\frac{1}{G}\sum_{i=1}^G \frac{1}{|o_i|}\sum_{t=1}^{|o_i|}
	\Bigl[\min\Bigl(\rho_{i,t}(\theta)\hat{A}_{i,t},\\
	&\text{clip}\bigl(\rho_{i,t}(\theta),1-\epsilon,1+\epsilon\bigr)\hat{A}_{i,t}\Bigr)
	-\beta\widehat{D}_{\mathrm{KL},i,t}\Bigr],
	\end{aligned}
	\label{eq:grpo_objective}
\end{equation}
\noindent where $\rho_{i,t}(\theta) = \frac{\pi_{\theta}(o_{i,t} \mid x, o_{i,<t})}{\pi_{\theta_{\text{old}}}(o_{i,t} \mid x, o_{i,<t})}$ is the probability ratio between the current policy and the old policy, $\text{clip}(\cdot, 1-\epsilon, 1+\epsilon)$ constrains the probability ratio to the interval $[1-\epsilon, 1+\epsilon]$ to prevent excessively large policy updates, $\epsilon$ is the clipping hyperparameter, and $\beta$ controls the strength of KL regularization.

This objective constrains policy updates without requiring a learned value model. Other GRPO-family implementations use batch-level or fixed-length token normalizers; we state the response-level normalizer explicitly because changing it can introduce a different length weighting.

\section{Method}

\begin{figure*}
	\centering
	\includegraphics[width=0.8\textwidth]{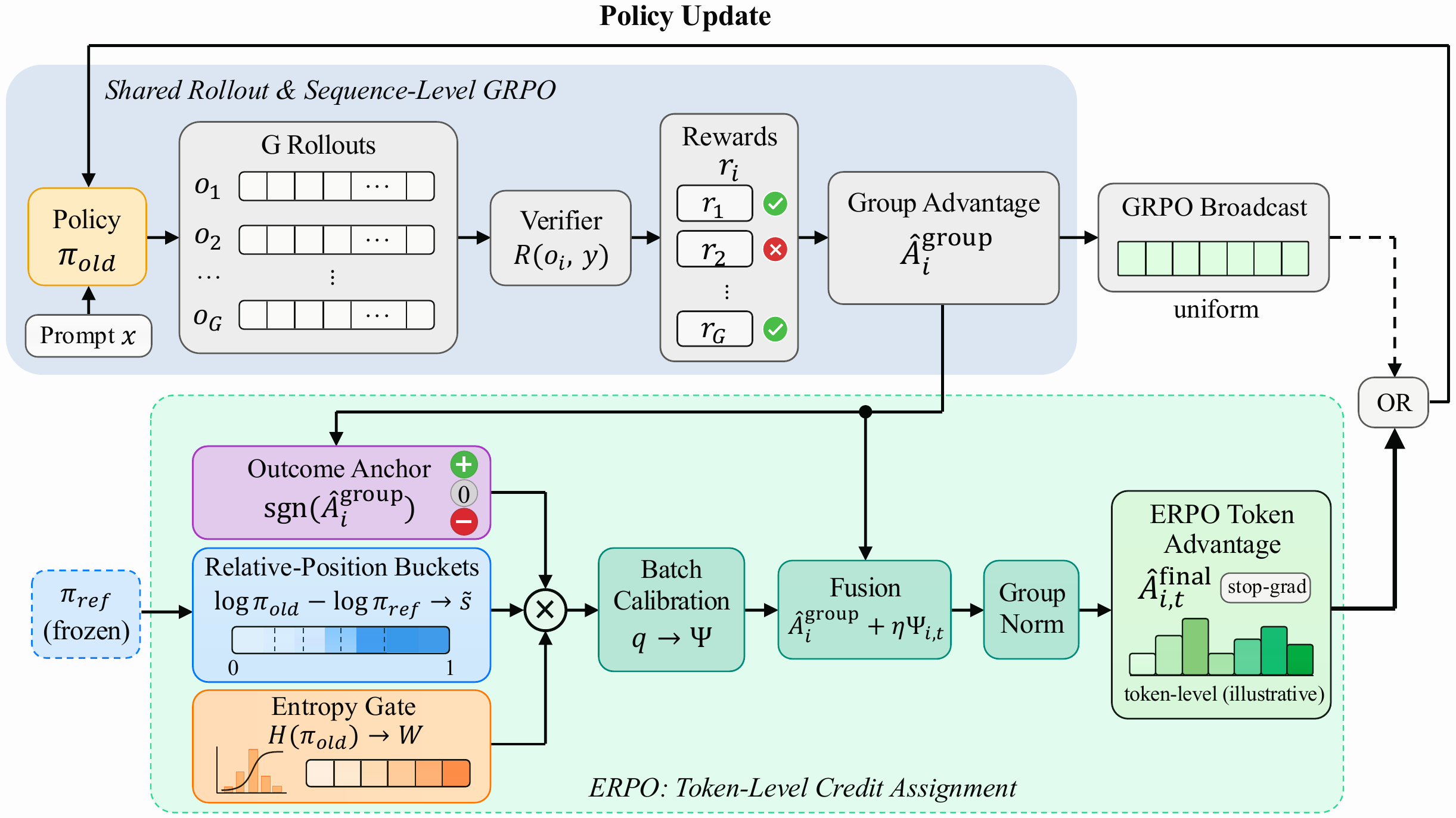}
	\caption{\textnormal{\textbf{Overview of ERPO within the GRPO pipeline.}
			GRPO (upper path) broadcasts each response's group-relative advantage $\hat A_i^{\mathrm{group}}$ to all its tokens.
			ERPO (lower path) multiplies the bucket-normalized policy-change signal $\tilde{s}_{i,t}$ by the entropy gate $W_{i,t}$ and the sign of the same advantage.
			After batch calibration, ERPO adds the scaled correction $\eta\Psi_{i,t}$ to $\hat A_i^{\mathrm{group}}$ and normalizes within the prompt group.
			The \emph{stop-grad} label indicates that ERPO holds the resulting token advantage $\hat A_{i,t}^{\mathrm{final}}$ fixed when differentiating the shared policy loss.}}
	\label{fig:erpo_architecture}
	\Description{A verifier scores responses sampled for one prompt. The group advantage branches to the GRPO broadcast, the ERPO outcome anchor, and the ERPO fusion step. GRPO assigns the same advantage to every token in a response. ERPO multiplies the outcome sign, entropy gate, and position-normalized policy change, rescales the product, and combines it with the group advantage. After group normalization, ERPO treats the token advantages as fixed weights for the policy update.}
\end{figure*}

To address heterogeneous token dynamics in reasoning chains, we propose Entropy-Regulated Policy Optimization (ERPO). ERPO refines a sequence-level group advantage into a token-level training signal using statistics computed from trajectories sampled by the behavior policy $\pi_{\mathrm{old}}$. All rollout-derived token statistics and constructed advantages are detached before the policy update. Figure \ref{fig:erpo_architecture} summarizes the method, and Algorithm \ref{alg:erpo} gives the training procedure.

\subsection{Rollout-Derived Token Signals}
We first define two detached token statistics computed during rollout. For a prompt group $\mathcal{G}$, let $\mathcal{M}_{\mathcal{G}}=\{(i,t):1\leq i\leq G,\ 1\leq t\leq |o_i|\}$ denote its active completion tokens. Unless stated otherwise, the ERPO token statistics below use population moments over the indicated active set. We write
\begin{equation}
	\operatorname{Std}_{\delta}(z;\mathcal{M})
	=\sqrt{\operatorname{Var}_{(i,t)\in\mathcal{M}}[z_{i,t}]+\delta}
\end{equation}
for a stabilized standard deviation, with $\delta=10^{-8}$ throughout the token-level normalizations.

\textbf{Uncertainty Estimation.} We employ the policy entropy $H_{i,t}$ as a statistical proxy for local predictive uncertainty and use it to construct a bounded token gate. Formally, for each token $o_{i,t}$, the entropy is defined as:
\begin{equation}
	H_{i,t} = -\sum_{v \in \mathcal{V}} \pi_{\mathrm{old}}(v\mid x, o_{i,<t}) \log \pi_{\mathrm{old}}(v\mid x, o_{i,<t}).
\end{equation}

This metric quantifies local predictive uncertainty. High entropy is used as a prioritization heuristic; it need not correspond to a semantic branch point in every trajectory.

\textbf{Reference-Relative Policy-Change Signal.}
For each sampled token, we define a reference-relative signal $s_{i,t}$ from the difference between its behavior-policy and frozen-reference log probabilities:
\begin{equation}
	s_{i,t} = \beta_{\text{progress}} \left( \log \pi_{\mathrm{old}}(o_{i,t} \mid x, o_{i,<t}) - \log \pi_{\text{ref}}(o_{i,t} \mid x, o_{i,<t}) \right),
\end{equation}
where $\beta_{\text{progress}}>0$ sets the numerical scale before bucket normalization. In the absence of the numerical stabilizer, the subsequent z-score removes any positive constant rescaling; we therefore fix this value rather than tune it as an independent source of performance. The signal records how the behavior policy changes the log probability of the sampled token relative to the reference. It is neither a token-correctness label nor a ground-truth process reward, which motivates the outcome-conditioned orientation introduced below.

\subsection{Entropy-Aware Gating}
While $H_{i,t}$ provides a raw measure of uncertainty, to effectively leverage this signal for optimization, we introduce an adaptive gating function $W_{i,t}$ that translates local predictive uncertainty into a relative importance weight. 
We perform intra-group entropy calibration to account for varying prompt-level uncertainty. Specifically, for each token $o_{i,t}$ in a prompt group $\mathcal{G}$, we compute the gated weight:
\begin{equation}
	W_{i,t} = w_{\min} + (1-w_{\min})\sigma\left(\gamma \cdot \frac{H_{i,t} - \mu_{H, \mathcal{G}}}{\operatorname{Std}_{\delta}(H;\mathcal{M}_{\mathcal{G}})}\right),
\end{equation}
where $\mu_{H, \mathcal{G}}$ is the mean over $\mathcal{M}_{\mathcal{G}}$; these group statistics are not moving averages across optimization steps. The sigmoid maps standardized entropy to a bounded weight, $\gamma$ controls its sharpness, and $w_{\min}=0.01$ prevents the gate from assigning an exact zero. This gate gives relatively uncertain positions more influence while avoiding an unbounded entropy bonus.

\subsection{Relative-Position Normalization}
As shown in Figure \ref{fig:entropy_map}, token statistics vary with position in a response. Directly pooling all positions can therefore mix early and late regions with different distributions. We use relative-position bucketing as an approximate control for this effect.

For one-based token index $t$, we define relative position as $\tau_{i,t}=(t-1)/|o_i|$. We partition $[0,1)$ into $K$ buckets $\mathcal{B}_0,\ldots,\mathcal{B}_{K-1}$. Within each prompt group $\mathcal{G}$, we normalize the reference-relative signal over active tokens assigned to the same bucket:
\begin{equation}
	\tilde{s}_{i,t} = \frac{s_{i,t} - \mu_{k, \mathcal{G}}}{\operatorname{Std}_{\delta}(s;\mathcal{M}_{k,\mathcal{G}})},
	\quad (i,t)\in\mathcal{M}_{k,\mathcal{G}},
\end{equation}
where $\mathcal{M}_{k,\mathcal{G}}=\mathcal{M}_{\mathcal{G}}\cap\mathcal{B}_k$ and $\mu_{k,\mathcal{G}}$ is the mean of $s$ over this set. Thus, the bucket statistics are computed across the $G$ responses to the same prompt, not across unrelated prompts. Comparing tokens at similar relative positions reduces positional scale differences, although it does not guarantee that two responses occupy semantically aligned reasoning stages.

\subsection{Final Advantage Synthesis}
We next combine the normalized policy-change signal with the sequence-level outcome advantage. Because policy change relative to the reference model is not evidence of correctness, using it without an outcome constraint could reinforce changes associated with unsuccessful trajectories.

We use the sign of the sequence-level advantage to orient the reference-relative deviations. Define the unscaled dense signal
\begin{equation}
	q_{i,t}=W_{i,t}\operatorname{sgn}(\hat{A}_i^{\text{group}})\tilde{s}_{i,t}.
\end{equation}
Let $\mathcal{M}_{\mathrm{batch}}$ contain all active tokens in the rollout generation batch. ERPO calibrates the dense component once over this set:
\begin{equation}
	\begin{aligned}
	\sigma_q&=\max\!\left(\sqrt{\operatorname{Var}_{\mathcal{M}_{\mathrm{batch}}}[q]},\delta\right),\\
	\Psi_{i,t}&=\sigma_{\text{target}}\frac{q_{i,t}}{\sigma_q}.
	\end{aligned}
\end{equation}
This batch-level calibration prevents the auxiliary component's empirical scale from varying directly with the raw log-probability difference. It is distinct from the prompt-group normalization below.

For $z_{i,t}=\hat{A}_{i}^{\text{group}}+\eta\Psi_{i,t}$, define
\begin{equation}
	\operatorname{Norm}_{\mathcal{G}}(z_{i,t})
	=\frac{z_{i,t}-\mu_{z,\mathcal{G}}}
	{\operatorname{Std}_{\delta}(z;\mathcal{M}_{\mathcal{G}})}.
\end{equation}
The final token advantage is
\begin{equation}
	\hat{A}_{i,t}^{\text{final}} = \operatorname{Norm}_{\mathcal{G}} \left( \hat{A}_{i}^{\text{group}} + \eta \Psi_{i,t} \right).
\end{equation}
Here, $\sigma_{\text{target}}$ controls the standard deviation of the dense component before mixing, and $\eta$ is its mixing coefficient. The outcome sign orients the reference-relative deviation but does not force every token correction to share the sign of the sequence advantage or identify whether an individual token is correct. The final normalization controls the empirical scale within each prompt group.
\begin{algorithm}[t]
	\caption{Entropy-Regulated Policy Optimization (ERPO)}
	\label{alg:erpo}
	
	\Require{Initial policy $\pi_\theta$, reference policy $\pi_{\text{ref}}$; learning rate $\alpha_{\text{lr}}$; group size $G$; number of buckets $K$; gating parameters $\gamma,w_{\min}$; scaling constants $\beta_{\text{progress}}, \eta, \sigma_{\text{target}}$.}
	\Ensure{Optimized policy parameters $\theta$.}
	
	\For{iteration $n = 1, \dots, N$}{
		Set $\pi_{\mathrm{old}}\leftarrow\pi_\theta$. For a batch from $\mathcal D$, sample $G$ responses per prompt with vLLM (temperature $1.2$, top-$p=0.95$) and save token log probabilities $\ell^{\mathrm{vLLM}}_{i,t}$\;
		Within each prompt group $\mathcal G$, compute rewards $r_i=R(o_i,y)$ and GRPO advantages $\hat{A}_i^{\text{group}}$\;
		
		\BlankLine
		\Comment{Full vocabulary, temperature $1.2$}
		\For{each response $o_i$ and valid token $t$}{
			$H_{i,t} \leftarrow -\sum_{v \in \mathcal{V}} \pi_{\mathrm{old}}(v \mid x, o_{i,<t}) \log \pi_{\mathrm{old}}(v \mid x, o_{i,<t})$\;
			$s_{i,t} \leftarrow \beta_{\text{progress}} \big( \log \pi_{\mathrm{old}}(o_{i,t} \mid \cdot) - \log \pi_{\text{ref}}(o_{i,t} \mid \cdot) \big)$\;
			$k \leftarrow \min\{K-1,\lfloor K(t-1)/|o_i|\rfloor\}$ and assign token $(i,t)$ to $\mathcal{B}_k$\;
		}
		
		\BlankLine
		\Comment{Prompt-group gating and normalization}
		$W_{i,t} \leftarrow w_{\min}+(1-w_{\min})\operatorname{Sigmoid}(\gamma\operatorname{Norm}_{\mathcal{G}}(H_{i,t}))$\;
		\For{each nonempty bucket $k$ in each prompt group $\mathcal G$}{
			$\tilde{s}_{i,t} \leftarrow \text{Norm}_{k,\mathcal{G}}(s_{i,t})$ for tokens in that bucket\;
		}
		
		\BlankLine
		\Comment{Outcome-Anchored Synthesis}
		$q_{i,t} \leftarrow W_{i,t} \cdot \text{sgn}(\hat{A}_i^{\text{group}}) \cdot \tilde{s}_{i,t}$\;
		$\Psi_{i,t}\leftarrow \sigma_{\text{target}}q_{i,t}/\max\{\operatorname{std}_{\mathrm{batch}}(q),\delta\}$ over all active tokens in the generation batch\;
		$\hat{A}_{i,t}^{\text{final}} \leftarrow \text{Norm}_{\mathcal{G}} \big( \hat{A}_i^{\text{group}} + \eta \cdot \Psi_{i,t} \big)$\;
		Stop the gradient through $\hat{A}_{i,t}^{\text{final}}$\;
		
		$c_{i,t}\leftarrow\min\{3,\exp(\log\pi_{\mathrm{old}}(o_{i,t}\mid\cdot)-\ell^{\mathrm{vLLM}}_{i,t})\}$\;
		Take one optimizer step on the batch mean of Eq.~\eqref{eq:grpo_objective} using $\hat{A}_{i,t}^{\text{final}}$ and weighting only the clipped surrogate by the detached $c_{i,t}$\;
	}
	\textbf{Return} $\theta$.
	
\end{algorithm}
\subsection{Optimization Interpretation and Scope}

ERPO computes entropy, reference-policy log probabilities, bucket statistics, gates, and final token advantages from trajectories sampled by the old policy. These quantities are detached during the subsequent policy update. Conditional on a sampled batch, ERPO therefore uses the same clipped surrogate loss as GRPO after replacing the broadcast sequence advantage with the constructed token advantage. This is an operational description of the implemented estimator, not an assertion that ERPO is exactly the gradient of a separate entropy-regularized objective.

The log ratio used to compute $s_{i,t}$ depends on the sampled token even after detachment. When it exceeds its bucket mean, the dense correction is positive for a response with positive group advantage and negative for one with negative group advantage. This outcome-dependent weighting can change the expected policy update. Its net effect on policy entropy or KL divergence also depends on the gate, final group normalization, and surrogate clipping.

For a non-degenerate group, the final z-score gives the active token advantages zero empirical mean and approximately unit empirical variance; the stabilizer handles degenerate groups. This controls their batchwise scale, but zero mean alone neither guarantees optimization stability nor prevents reward hacking. In particular, ERPO inherits the correctness and coverage limitations of the verifiable reward. In the matched 3B runs, all $G=8$ responses received the same reward in 46.45\% of ERPO prompt groups and 45.25\% of GRPO groups. In such groups, the zero group advantage suppresses ERPO's dense correction, so only the KL term can contribute a gradient. Finally, the normalization is token weighted, so unequal response lengths can affect the group statistics. We treat these properties as explicit limitations and evaluate optimization stability, response length, and component contributions empirically.

\section{Experiments}
\begin{table*}
	\centering
	\small 
	\rmfamily
	\caption{\textnormal{Overall performance results. Acc (\%) and Fmt (\%) represent sample accuracy and the boxed-answer rate, respectively. For the 3B scale, ERPO–GRPO results are reported as mean ± sample standard deviation (based on seeds 42, 43, and 44); all other results are from single runs. The highest mean in each Acc column is bolded, and the second-highest distinct Acc mean is underlined.}}
	\label{tab:qwen_reordered_results}
	\renewcommand{\arraystretch}{1.2}
	\setlength{\tabcolsep}{3.2pt}
	
	\begin{tabular}{c cc cc cc cc | cc}
		\toprule
		& \multicolumn{2}{c}{\textbf{AMC23}} & \multicolumn{2}{c}{\textbf{Minerva}} & \multicolumn{2}{c}{\textbf{AIME24}} & \multicolumn{2}{c}{\textbf{AIME25}} & \multicolumn{2}{c}{\textbf{Average}} \\
		\cmidrule(lr){2-3} \cmidrule(lr){4-5} \cmidrule(lr){6-7} \cmidrule(lr){8-9} \cmidrule(lr){10-11}
		\textbf{Model} & {Acc} & {Fmt} & {Acc} & {Fmt} & {Acc} & {Fmt} & {Acc} & {Fmt} & \textbf{Acc} & \textbf{Fmt} \\
		\midrule
		
		\rowcolor{TableGroupGray}
		\multicolumn{11}{c}{\textit{1.5B Scale}} \\
		Qwen2.5-1.5B$_{\text{Base}}$ & 1.48 & 24.14 & 0.08 & 22.66 & 0.31 & 26.35 & 0.00 & 25.52 & 0.47 & 24.67 \\
		Qwen2.5-1.5B$_{\text{SFT}}$ & 8.13 & 98.28 & 1.41 & 84.84 & 0.83 & 95.42 & 0.42 & 93.13 & 2.70 & 92.92 \\
		Qwen2.5-1.5B$_{\text{GRPO}}$ & \underline{24.77} & 98.12 & \underline{4.77} & 98.59 & \underline{2.71} & 91.25 & \underline{0.73} & 97.60 & \underline{8.24} & 96.39 \\
		\rowcolor{TableMethodBlue}
		\textbf{Qwen2.5-1.5B$_{\text{ERPO}}$} & \textbf{27.89} & 95.94 & \textbf{6.17} & 97.11 & \textbf{3.12} & 88.44 & \textbf{1.77} & 96.25 & \textbf{9.74} & 94.43 \\
		
		\midrule
		\rowcolor{TableGroupGray}
		\multicolumn{11}{c}{\textit{3B Scale}} \\
		Qwen2.5-3B$_{\text{Base}}$ & 16.33 & 72.03 & 3.52 & 58.67 & 2.08 & 78.54 & 0.94 & 79.48 & 5.72 & 72.18 \\
		Qwen2.5-3B$_{\text{SFT}}$ & 10.63 & 99.38 & 2.03 & 74.69 & 0.83 & 95.42 & 0.83 & 95.63 & 3.58 & 91.28 \\
	Qwen2.5-3B$_{\text{GRPO}}$ & \underline{34.66}\,{\scriptsize$\pm$0.60} & 97.37\,{\scriptsize$\pm$0.39} & \underline{11.07}\,{\scriptsize$\pm$0.78} & 98.72\,{\scriptsize$\pm$0.20} & \underline{4.58}\,{\scriptsize$\pm$0.75} & 92.88\,{\scriptsize$\pm$0.26} & \underline{1.81}\,{\scriptsize$\pm$0.51} & 97.74\,{\scriptsize$\pm$0.16} & \underline{13.03}\,{\scriptsize$\pm$0.32} & 96.68\,{\scriptsize$\pm$0.11} \\
		Qwen2.5-3B$_{\text{Dr.~GRPO}}$ & 27.11 & 71.80 & 10.31 & 98.75 & 3.85 & 84.58 & 0.83 & 55.83 & 10.53 & 77.74 \\
		Qwen2.5-3B$_{\text{DAPO-style}}$ & 30.31 & 72.19 & 9.61 & 98.59 & 3.96 & 77.81 & 0.62 & 50.10 & 11.13 & 74.67 \\
		\rowcolor{TableMethodBlue}
	\textbf{Qwen2.5-3B$_{\text{ERPO}}$} & \textbf{35.47}\,{\scriptsize$\pm$0.43} & 96.74\,{\scriptsize$\pm$0.25} & \textbf{11.72}\,{\scriptsize$\pm$0.43} & 98.75\,{\scriptsize$\pm$0.16} & \textbf{5.38}\,{\scriptsize$\pm$0.30} & 92.22\,{\scriptsize$\pm$0.67} & \textbf{2.05}\,{\scriptsize$\pm$0.22} & 97.53\,{\scriptsize$\pm$0.42} & \textbf{13.65}\,{\scriptsize$\pm$0.14} & 96.31\,{\scriptsize$\pm$0.25} \\
		
		\midrule
		\rowcolor{TableGroupGray}
		\multicolumn{11}{c}{\textit{7B Scale}} \\
		Qwen2.5-7B$_{\text{Base}}$ & 24.77 & 84.30 & 6.80 & 77.81 & 4.48 & 83.12 & 1.67 & 82.40 & 9.43 & 81.91 \\
		Qwen2.5-7B$_{\text{SFT}}$ & 17.03 & 96.88 & 4.84 & 71.88 & 1.46 & 88.75 & 1.46 & 94.79 & 6.20 & 88.08 \\
		Qwen2.5-7B$_{\text{GRPO}}$ & \underline{47.34} & 99.06 & \underline{18.98} & 99.77 & \underline{9.79} & 96.35 & \underline{5.73} & 97.29 & \underline{20.46} & 98.12 \\
		\rowcolor{TableMethodBlue}
		\textbf{Qwen2.5-7B$_{\text{ERPO}}$} & \textbf{50.23} & 98.75 & \textbf{20.00} & 99.38 & \textbf{11.88} & 96.35 & \textbf{6.77} & 98.23 & \textbf{22.22} & 98.18 \\
		
		\bottomrule
	\end{tabular}
\end{table*}
\subsection{Experimental Setup}
\textbf{Datasets and Evaluation Benchmarks.} 
For reinforcement learning, we use problems at difficulty levels 3--5 from MATH \citep{hendrycks2021measuring}. We retain prompts of at most 204 tokens after applying the chat template and discard examples whose reference solution does not contain an extractable boxed answer. We evaluate out-of-distribution mathematical reasoning on AMC23\footnote{\url{https://huggingface.co/datasets/math-ai/amc23}}, AIME24\footnote{\url{https://huggingface.co/datasets/math-ai/aime24}}, AIME25\footnote{\url{https://huggingface.co/datasets/math-ai/aime25}}, and Minerva \citep{lewkowycz2022solving}. For the base and RL-trained models, we sample 32 responses per test problem at temperature 1.0 and top-$p=0.95$. A response is correct if either the rule-based verifier or \texttt{math-verify}~\citep{kydlicek2026mathverify} accepts its last boxed answer. We compute $pass@k$ as the fraction of size-$k$ subsets of the 32 responses containing at least one correct answer, averaged over problems.

\textbf{Model Configurations.} 
We implement ERPO on Qwen2.5 models \citep{qwen2025qwen25} with 1.5B, 3B, and 7B parameters. We use LoRA \citep{hu2022lora} with rank $r=32$ and $\alpha_{\mathrm{LoRA}}=64$, targeting the attention projections and the gate, up, and down projections in each MLP block. We set $K=10$, $\beta_{\text{progress}}=0.1$, $\gamma=5$, $\eta=0.2$, and $\sigma_{\text{target}}=0.15$.

\textbf{Training Specifications.} 
We train each model for one epoch using TRL~\citep{vonwerra2020trl} with colocated vLLM~\citep{kwon2023pagedattention} generation on one RTX PRO 6000 Blackwell GPU. The maximum completion length is 2,048 tokens; with the 204-token prompt filter, the maximum model context used by generation is 2,252 tokens. Each optimizer update contains 64 sampled completions, corresponding to eight distinct prompts with $G=8$ responses per prompt. The per-device completion batch size and gradient-accumulation steps are $(16,4)$ for 1.5B and 3B and $(8,8)$ for 7B. We use a learning rate of $5\times10^{-6}$, a 0.1 warmup ratio followed by linear decay, an 8-bit AdamW optimizer with weight decay 0.001, BF16 model computation, and a KL coefficient of 0.001. We use symmetric policy-ratio clipping with $\epsilon=0.2$, perform one optimizer update per rollout batch ($\mu=1$), and clip the gradient norm at 1.0. The sampling temperature is 1.2 with top-$p=0.95$. All compared policy-optimization variants use the same data, rollout budget, reward function, and generation settings.
\subsection{Baselines}
\begin{table*}[t]
	\centering
	\caption{\textnormal{Comparison of $pass@k$ (\%) across benchmarks. For the 3B scale, the ERPO and GRPO figures represent the mean results from random seeds 42, 43, and 44, while the other figures are based on single runs. Across all parameter scales, the highest value in each column is bolded, and the second-highest value is underlined.}}
	\label{tab:qwen_pass_k_corrected}
	\renewcommand{\arraystretch}{1.2}
	\setlength{\tabcolsep}{1.5pt}
	
	\resizebox{\textwidth}{!}{ 
		\rmfamily
		\begin{tabular}{l | cccc | cccc | cccc | cccc}
			\toprule
			& \multicolumn{4}{c|}{\textbf{AMC23}} & \multicolumn{4}{c|}{\textbf{Minerva}} & \multicolumn{4}{c|}{\textbf{AIME24}} & \multicolumn{4}{c}{\textbf{AIME25}} \\
			\textbf{Model} & @2 & @4 & @8 & @16 & @2 & @4 & @8 & @16 & @2 & @4 & @8 & @16 & @2 & @4 & @8 & @16 \\
			\midrule
			
			\rowcolor{TableGroupGray}
			\multicolumn{17}{c}{\textit{1.5B Scale}} \\
			Qwen2.5-1.5B$_{\text{Base}}$ & 2.89 & 5.48 & 9.98 & 17.32 & 0.16 & 0.31 & 0.62 & 1.25 & 0.62 & 1.21 & 2.31 & 4.19 & 0.00 & 0.00 & 0.00 & 0.00 \\
			Qwen2.5-1.5B$_{\text{SFT}}$  & 14.90 & 25.41 & 39.17 & 55.00 & 2.48 & 3.96 & 5.66 & 7.50 & 1.67 & 3.33 & 6.67 & 13.33 & 0.83 & 1.67 & 3.33 & 6.67 \\
			Qwen2.5-1.5B$_{\text{GRPO}}$ & \underline{34.69} & \underline{45.36} & \underline{56.09} & \underline{65.64} & \underline{7.28} & \underline{10.12} & \underline{13.51} & \underline{18.19} & \underline{4.65} & \underline{7.35} & \underline{10.98} & \underline{16.12} & \underline{1.42} & \underline{2.69} & \underline{4.85} & \underline{8.16} \\
			\rowcolor{TableMethodBlue}
			\textbf{Qwen2.5-1.5B$_{\text{ERPO}}$} & \textbf{38.72} & \textbf{49.86} & \textbf{60.87} & \textbf{71.13} & \textbf{8.81} & \textbf{11.65} & \textbf{14.99} & \textbf{19.25} & \textbf{5.39} & \textbf{8.53} & \textbf{12.61} & \textbf{18.38} & \textbf{3.33} & \textbf{5.90} & \textbf{9.47} & \textbf{13.33} \\
			
			\midrule
			\rowcolor{TableGroupGray}
			\multicolumn{17}{c}{\textit{3B Scale}} \\
			Qwen2.5-3B$_{\text{Base}}$    & 27.45 & 41.47 & 55.90 & 68.84 & 6.10 & 9.64 & 13.71 & 18.12 & 3.82 & 6.49 & 9.82 & 13.09 & 1.83 & 3.48 & 6.33 & 10.69 \\
			Qwen2.5-3B$_{\text{SFT}}$     & 18.31 & 28.64 & 40.35 & 52.50 & 3.77 & 6.56 & 10.38 & 15.00 & 1.64 & 3.17 & 5.89 & 10.00 & 1.67 & 3.33 & 6.67 & 13.33 \\
			Qwen2.5-3B$_{\text{GRPO}}$ & \underline{47.28} & \textbf{59.76} & \underline{70.15} & \underline{78.27} & \underline{16.12} & \underline{21.68} & \underline{27.74} & \underline{33.85} & \underline{7.44} & \underline{10.89} & 15.13 & 20.94 & \underline{3.46} & \underline{6.40} & \underline{11.11} & \underline{17.76} \\
			Qwen2.5-3B$_{\text{Dr.~GRPO}}$ & 37.14 & 48.50 & 59.87 & 71.26 & 15.09 & 19.83 & 24.73 & 29.93 & 6.25 & 8.98 & 11.95 & 16.29 & 1.64 & 3.17 & 5.95 & 10.48 \\
			Qwen2.5-3B$_{\text{DAPO-style}}$ & 41.60 & 53.73 & 64.11 & 70.95 & 14.23 & 19.11 & 24.64 & 31.13 & 6.88 & 10.88 & \underline{15.67} & \underline{21.40} & 1.24 & 2.46 & 4.81 & 9.19 \\
			\rowcolor{TableMethodBlue}
			\textbf{Qwen2.5-3B$_{\text{ERPO}}$} & \textbf{47.65} & \underline{59.68} & \textbf{70.31} & \textbf{79.28} & \textbf{17.04} & \textbf{23.16} & \textbf{29.96} & \textbf{36.84} & \textbf{8.60} & \textbf{12.48} & \textbf{17.36} & \textbf{23.91} & \textbf{3.88} & \textbf{7.03} & \textbf{11.91} & \textbf{18.69} \\
			
			\midrule
			\rowcolor{TableGroupGray}
			\multicolumn{17}{c}{\textit{7B Scale}} \\
			Qwen2.5-7B$_{\text{Base}}$    & 38.91 & 53.35 & 64.72 & 73.51 & 11.76 & 18.36 & 25.42 & 31.88 & 7.96 & 13.05 & 19.55 & 26.96 & 3.27 & 6.31 & 11.72 & 20.16 \\
			Qwen2.5-7B$_{\text{SFT}}$     & 28.56 & 43.31 & 59.20 & 70.00 & 8.17 & 12.31 & 16.47 & 20.00 & 2.86 & 5.50 & 10.11 & 16.67 & 2.83 & 5.36 & 9.67 & 16.67 \\
			Qwen2.5-7B$_{\text{GRPO}}$ & \underline{59.77} & \underline{70.82} & \underline{78.53} & \underline{83.73} & \underline{24.95} & \underline{31.65} & \underline{37.75} & \underline{42.81} & \underline{14.39} & \underline{19.72} & \underline{25.45} & \textbf{31.26} & \underline{10.04} & \underline{16.00} & \textbf{22.85} & \textbf{30.38} \\
			\rowcolor{TableMethodBlue}
			\textbf{Qwen2.5-7B$_{\text{ERPO}}$} & \textbf{63.38} & \textbf{73.33} & \textbf{80.47} & \textbf{86.08} & \textbf{26.54} & \textbf{32.96} & \textbf{38.98} & \textbf{43.83} & \textbf{16.18} & \textbf{20.75} & \textbf{25.50} & \underline{30.62} & \textbf{11.50} & \textbf{17.33} & \underline{22.49} & \underline{25.61} \\
			
			\bottomrule
		\end{tabular}
	} 
\end{table*}
To evaluate the effectiveness of ERPO, we benchmark our approach against the following baseline configurations:

\textbf{Base Model.} The corresponding pretrained Qwen2.5 model without task-specific fine-tuning.

\textbf{SFT Model.} A checkpoint obtained by supervised fine-tuning on the same MATH data.

\textbf{GRPO.} The primary baseline uses the clipped objective in Eq.~\ref{eq:grpo_objective}, within-group reward standardization, and per-response token averaging under the same rollout and reward configuration as ERPO.

\textbf{Dr.~GRPO.} This matched variant removes within-group reward standard-deviation scaling and divides the summed token loss by the fixed maximum completion length.

\textbf{DAPO-style.} This variant changes only the GRPO loss normalization, dividing the total token loss by the number of active completion tokens in the global accumulated batch.

\textbf{Component Ablations.} On Qwen2.5-3B, \emph{w/o entropy gate} sets $W_{i,t}=1$, \emph{w/o relative-position buckets} normalizes the implicit signal once over all active tokens in the prompt group, and \emph{w/o outcome anchor} replaces $\operatorname{sgn}(\hat{A}_i^{\text{group}})$ with one. All other settings, including the sampled-completion budget and maximum completion length, are unchanged.

\begin{table}[t]
	\centering
	\small
	\rmfamily
	\caption{\textnormal{Ablation of the 3B component at seed 42. The values represent sample accuracy (\%).}}
	\label{tab:erpo_ablation_oracle}
	\renewcommand{\arraystretch}{1.12}
	\setlength{\tabcolsep}{3.2pt}
	\resizebox{\columnwidth}{!}{%
	\begin{tabular}{lccccc}
		\toprule
		\textbf{Method} & \textbf{AMC23} & \textbf{Minerva} & \textbf{AIME24} & \textbf{AIME25} & \textbf{Avg.} \\
		\midrule
		GRPO & 34.14 & 11.02 & 3.96 & 1.56 & 12.67 \\
		ERPO & \textbf{35.94} & \textbf{11.64} & \textbf{5.21} & \textbf{1.98} & \textbf{13.69} \\
		\quad w/o entropy gate & 34.14 & 10.94 & 3.85 & 1.56 & 12.62 \\
		\quad w/o position buckets & 35.62 & 11.48 & 5.00 & 1.56 & 13.42 \\
		\quad w/o outcome anchor & 0.00 & -- & -- & -- & -- \\
		\bottomrule
	\end{tabular}%
	}
\end{table}

\subsection{Overall Performance}
As shown in Table~\ref{tab:qwen_reordered_results}, ERPO improves the four-benchmark average sample accuracy over matched GRPO from 8.24 to 9.74 at 1.5B, from 13.03 to 13.65 at 3B, and from 20.46 to 22.22 at 7B. The gains vary across benchmarks and model scales. Boxed-answer rates remain high for ERPO and GRPO, although neither method uniformly dominates this formatting metric.

Some SFT models underperform their corresponding base models on the harder evaluation sets. One possible explanation is a distribution mismatch between MATH supervision and the competition benchmarks, but the present experiments do not isolate this cause. We therefore treat SFT as a reference point rather than evidence that supervised fine-tuning generally reduces reasoning ability.
\subsection{Multi-Sample Scaling Analysis} 
As shown in Table~\ref{tab:qwen_pass_k_corrected}, the probability that at least one of the $k$ responses to a given problem is correct increases with $k$. ERPO improves over matched GRPO for most scale--benchmark--$k$ combinations, with especially clear gains at 1.5B and on AMC23. The comparison therefore supports improved multi-sample coverage relative to matched GRPO.

\begin{figure*}[t]
	\centering
	\includegraphics[width=\textwidth]{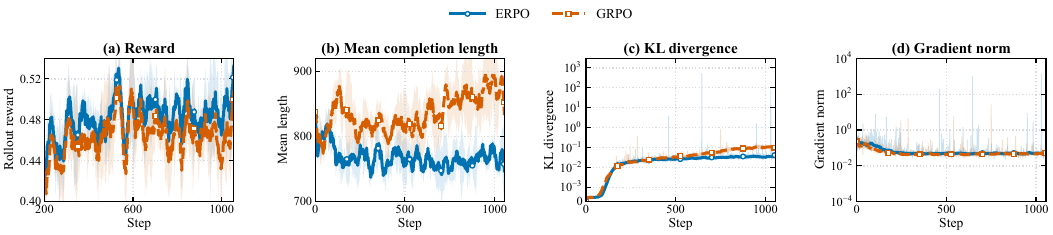}
	\caption{\textnormal{\textbf{Matched 3B training dynamics.}
			Thick curves are the across-seed means of 25-step smoothed training logs for seeds 42, 43, and 44; shaded regions denote $\pm$ one sample standard deviation across seeds.
			For the log-scale KL-divergence and gradient-norm panels, faint curves additionally show the unsmoothed across-seed means.}}
	\label{fig:training_dynamics_3b}
	\Description{Four line plots compare ERPO and GRPO over training steps. Late in training, ERPO generally has higher rollout reward, shorter completions, and lower smoothed KL divergence. The smoothed gradient norms largely overlap. The unsmoothed KL and gradient-norm traces have isolated spikes.}
\end{figure*}

Figure \ref{fig:training_dynamics_3b} provides training-side diagnostics for the central three-seed 3B comparison. Over the final 100 updates, ERPO obtains a mean rollout reward of 0.492 versus 0.465 for GRPO, while its sampled completions average 770 rather than 878 tokens. This paired pattern appears in all three seeds. The combination of higher reward with shorter average completions suggests that ERPO improves reasoning efficiency during training, achieving stronger outcomes without relying on longer sampled responses. It concerns the training distribution and sampling configuration: under the fixed benchmark evaluation used in Figure \ref{fig:baseline_behavior}, ERPO and GRPO instead remain in a similar response-length regime. The typical late-stage KL is lower for ERPO, and the median gradient-norm trajectories are similar. ERPO's unsmoothed three-seed means peaked at 520.62 for KL (step 646) and 1,387.21 for gradient norm (step 1,030). Seed 42 drove both peaks. One update later, the means fell to 0.026 and 0.057, respectively. GRPO's corresponding maxima were 0.80 and 27.74.

\subsection{Output Stability of Matched Baselines}

\begin{figure*}[t]
	\centering
	\includegraphics[width=\textwidth]{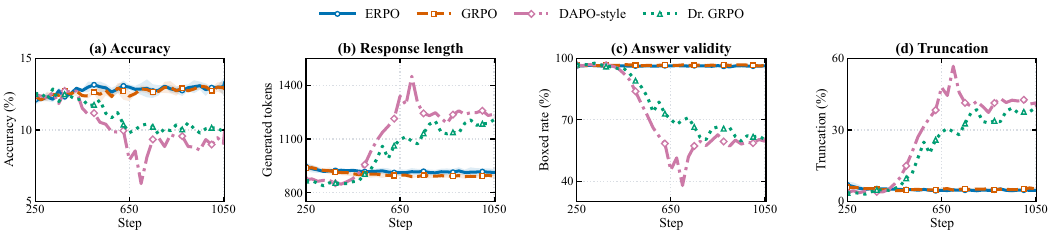}
	\caption{\textnormal{\textbf{Output behavior of matched 3B baselines.}
			All panels report unsmoothed four-benchmark macro-averages over the common checkpoint range.
			ERPO and GRPO show the mean and $\pm$ one sample standard deviation over seeds 42, 43, and 44; the DAPO-style and Dr.~GRPO curves are single-seed diagnostics.}}
	\label{fig:baseline_behavior}
	\Description{Four line plots track accuracy, response length, boxed-answer rate, and truncation. ERPO and GRPO remain close throughout. As training progresses, DAPO-style and Dr. GRPO produce longer responses, more truncations, fewer boxed answers, and lower accuracy.}
\end{figure*}

Figure \ref{fig:baseline_behavior} shows that ERPO and GRPO maintain evaluation truncation rates below 8\% and boxed-answer rates near 96\%. In the seed-42 runs, training truncation rose from 6.8\% to 47.5\% for DAPO-style and from 6.6\% to 46.0\% for Dr.~GRPO between steps 201--300 and the final 100 updates. Over the same windows, normally terminated responses grew shorter on average (672 to 494 tokens and 668 to 497 tokens, respectively). Truncated rollouts still contribute to the loss, and the verifier scores the last complete boxed answer. At the final checkpoint, about 40\% of benchmark responses from each variant reached the 2,048-token limit. Across the four benchmarks, accuracy on truncated responses averaged 1.7\% for DAPO-style and 0.8\% for Dr.~GRPO, compared with about 14\% on normally terminated responses for both.

\subsection{Component Analysis}
Table \ref{tab:erpo_ablation_oracle} isolates the three components on Qwen2.5-3B. Full ERPO reaches 13.69\% average accuracy, 1.02 points above GRPO. Removing the entropy gate reduces the average to 12.62\% and eliminates this gain on every benchmark; removing relative-position buckets produces a smaller 0.27-point decrease to 13.42\%. The gate therefore accounts for the largest measured component gain, while position-aware normalization provides an incremental refinement.

Without the outcome anchor, the run yields 0\% on AMC23 and no usable results on the remaining benchmarks. This failure is consistent with the role of $\operatorname{sgn}(\hat A_i^{\mathrm{group}})$ in coupling the dense signal to verifier feedback. Because $\tilde{s}_{i,t}$ measures policy change rather than correctness, the unsigned variant treats positive-, negative-, and zero-advantage trajectories alike. Crucially, when all responses receive the same reward, standard ERPO sets $q_{i,t}=0$, whereas the unsigned variant can retain $q_{i,t}=W_{i,t}\tilde{s}_{i,t}\neq0$. Group normalization can then rescale this reward-free variation to an approximately unit-scale advantage, enabling updates unsupported by any verifier preference. Repeatedly reinforcing deviations merely because they differ from the reference creates a positive-feedback mechanism that plausibly explains the observed drift and collapse. 
\subsection{Efficiency and Computational Overhead}

The matched 3B runs took 17.66 hours for ERPO and 17.71 for GRPO (Table~\ref{tab:training_efficiency}). Mean step times were 57.69 and 57.87 seconds, respectively. ERPO's mean rollout length was $8.1\%$ lower and its logged token count was $7.3\%$ lower. Throughput was 913.0 logged tokens/s for ERPO and 982.5 for GRPO. Differences in rollout length prevent these timings from isolating the cost of ERPO's token-level operations. 

\begin{table}[t]
	\centering
	\small
	\rmfamily
	\caption{\textnormal{End-to-end Qwen2.5-3B training cost. Runtime, optimizer-step time, cumulative logged training tokens, and rollout length are mean $\pm$ sample standard deviation over matched seeds 42/43/44 and 1,055 updates. }}
	\label{tab:training_efficiency}
	\renewcommand{\arraystretch}{1.12}
	\setlength{\tabcolsep}{2.7pt}
	\resizebox{\columnwidth}{!}{%
		\begin{tabular}{lccccc}
			\toprule
			\textbf{Method} & \makecell{\textbf{Runtime}\\\textbf{(h)}} & \makecell{\textbf{Step time}\\\textbf{(s)}} & \makecell{\textbf{Tokens}\\\textbf{(M)}} & \makecell{\textbf{Mean}\\\textbf{length}} & \makecell{\textbf{GPU memory}\\\textbf{(GiB)}} \\
			\midrule
			GRPO & $17.71\pm0.03$ & $57.87\pm0.11$ & $62.62\pm1.47$ & $835.6\pm21.8$ & 89.90 \\
			\rowcolor{TableMethodBlue}
			\textbf{ERPO} & \boldmath$17.66\pm0.10$ & \boldmath$57.69\pm0.34$ & \boldmath$58.05\pm0.60$ & \boldmath$767.9\pm8.8$ & 89.90 \\
			\bottomrule
		\end{tabular}%
	}
\end{table}
\section{Limitations}

Our empirical scope is mathematical RLVR with a binary final-answer verifier, so transfer to code execution, search or recommendation agents, or other web-facing verifiable tasks is not established. The verifier has incomplete coverage of mathematically equivalent answers and can produce false-negative rewards; ERPO inherits this supervision error because its dense signal is anchored only by the observed sequence outcome. Entropy and reference-relative log-probability change are operational proxies rather than ground-truth process rewards, and relative position does not guarantee alignment of reasoning stages across responses. Full-vocabulary entropy also adds computation and memory traffic, with cost depending on model and system implementation. ERPO receives no reward-driven gradient from prompt groups with identical rewards, and token-weighted group statistics depend on response length.
\section{Conclusion}

We presented ERPO, a method that uses entropy to refine sequence-level RLVR advantages into token-level training signals. ERPO combines an uncertainty gate, relative-position normalization of policy progress, and an outcome anchor without requiring a learned process-reward model.

Across three model scales and four mathematical benchmarks, the experiments evaluate the accuracy--length trade-off, optimization dynamics, component contributions, and end-to-end cost relative to a matched GRPO baseline. The current scope is limited to mathematical RLVR, and entropy and response length remain proxy measurements rather than direct evidence of semantic diversity or redundancy. When all responses to a prompt receive the same reward, their reward-based advantages are zero, but the KL term can still contribute a gradient. Future work should address zero-variance groups, test non-mathematical verifiable tasks, and study normalization schemes that reduce token-length weighting.

\section{Ethical Considerations}
ERPO learns from a verifier that checks the final answer. A response can therefore receive a positive reward even when an intermediate step is invalid. Conversely, a verifier can reject a mathematically equivalent answer and penalize a sound derivation. In an educational application, either error could affect the explanations shown to learners. The benchmark scores reported here measure final-answer accuracy. Assessing the reliability of explanations would require checking the reasoning steps themselves.

\bibliographystyle{ACM-Reference-Format}
\bibliography{sample-base}

\end{document}